\documentclass[10pt,twocolumn,letterpaper]{article}

\usepackage[pagenumbers]{cvpr} 
 \usepackage{multirow}

\usepackage{multirow} 

\definecolor{cvprblue}{rgb}{0.21,0.49,0.74}
\usepackage[pagebackref,breaklinks,colorlinks,allcolors=cvprblue]{hyperref}

\def\confName{CVPR}
\def\confYear{2027}

\title{IDTCount: Counting Every Instance in Open Vocabulary Video Counting}

\author{Joana Konadu Owusu, Shivanand Venkanna Sheshappanavar\\
Geometric Intelligence Research Lab, Dept. of Electrical Engineering and Computer Science\\
University of Wyoming, 1000 E University Ave, Laramie, WY 82071\\
{\tt\small \{jowusu1, ssheshap\}@uwyo.edu}
}

\begin{document}
\maketitle
\begin{abstract}
Open-vocabulary video counting counts distinct objects that match a text query across a video. The core challenge is maintaining consistent object identities across scene changes to prevent double counting or merged identities. To address this, we introduce IDTCount, an identity-driven temporal counting framework that maintains and validates persistent object identities across a video. It recovers lost identities during tracking and uses evidence from completed trajectories to resolve duplicate identities and suppress spurious ones. Across five counting benchmarks spanning long-tailed scenes, dense crowds, and long videos, IDTCount outperforms CountVid and SAM3Count, reducing CountVid's counting error (MAE) by 62.5\% on BURST-Count and 78.2\% on TAO-Count, and SAM3Count's by 9.6\% and 27.6\%. Furthermore, on three video instance segmentation benchmarks, IDTCount improves identity tracking over CountVid by up to 8.8\% in AssA, 55.8\% in HOTA, and 78.1\% in IDF1. Code is available at \href{https://github.com/Joan947/IDTCount}{https://github.com/Joan947/IDTCount}.
\end{abstract}
      
\section{Introduction}
\label{sec:intro}

\begin{figure}[t]
  \centering
  \includegraphics[width=0.99\linewidth]{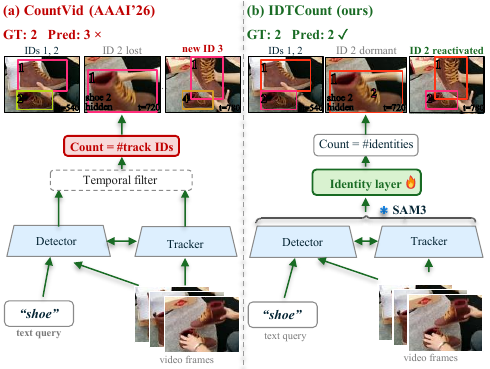}
  \caption{Given a text query (``shoe'') (a) CountVid~\cite{amini2025countvid} equates the final count with the number of generated track IDs. When an object is lost during occlusion and re-detected, it receives a new ID; here, a shoe is counted twice due to occlusion and re-entry. (b) Our IDTCount maintains a persistent identity for each distinct instance across occlusions, overlaps, and reappearances, yielding an accurate count. } \label{fig:teaser}
\end{figure}

Counting objects specified by natural language queries is known as open-vocabulary counting (OVC). This approach enables zero-shot enumeration without category-specific training~\cite{owusu2026objectcountingmodalitiestaxonomies}. While prompt-driven counting has rapidly matured for static images~\cite{amini2024countgd,AminiNaieni_2026_CountGD,Qian_2025_T2ICount}, extending OVC to video streams~\cite{amini2025countvid,Clark_2026_CVPR} is essential for dynamic real-world domains such as wildlife conservation~\cite{T_2026_FlockCounter,J_2022_Caltech}, cellular microscopy~\cite{Morelli_2021_Automating,S_2024_CoNIC}, crowd analytics~\cite{Wang_2020_NWPU,McCarthy_2025_Video}, traffic monitoring~\cite{XY_2024_SCU,GuerreroGomezOlmedo_2015_Extremely}, and aerial surveying~\cite{Tan_2026_Dense,Gao_2020_Counting_2}. Open-vocabulary counting in videos estimates the number of distinct physical instances that match a text query across a sequence. This requires deciding whether an object seen after an occlusion, a missed detection, or a re-entry is one we have already counted. (Figure~\ref{fig:teaser}a).

CountVid~\cite{amini2025countvid}, the pioneer in this domain, pairs its open-vocabulary detector with a video tracker ~\cite{ravi2024sam2} and outputs the number of generated tracks as the final count. Such tracking-based counters depend on the underlying tracker to maintain identity continuity across the video. These trackers rely on short-term association~\cite{weng2020gnn3dmot,pang2021qdtrack} and can lose an object during occlusion or scene exit, assigning it a new ID when it reappears (Figure~\ref{fig:teaser}a). As a result, CountVid degrades under heavy overlaps, dense scenes, and prolonged occlusions. SAM3Count~\cite{Owusu_2026_CVPR} attempts to address this by building a re-identification tracker on top of SAM3's tracker to recover lost tracks due to occlusion. However, it inherits SAM3's detection and tracking errors, degrading performance under heavy overlaps, prolonged occlusions, and dense scenes.  Moreover, recovering temporal correspondence alone does not guarantee an accurate count. Residual duplicate trajectories and false detections still introduce spurious identities that inflate the final prediction. These challenges highlight the need for an open vocabulary video counting framework that can recover previously seen objects while verifying whether each accumulated identity represents a distinct, valid instance.
To address this, we introduce \textbf{IDTCount}, an identity-driven temporal counting framework built on frozen observations that maintains and validates persistent object identities across a video. It recovers lost identities during tracking and leverages evidence from completed trajectories to filter spurious tracks and resolve duplicate identities. The final count prediction is therefore the number of retained persistent identities (Figure~\ref{fig:teaser}b). Our main contributions are:
\begin{itemize}
    \item We formulate open-vocabulary video counting as persistent identity management, addressing both identity recovery during tracking and validation of the accumulated identities for counting.
    \item We introduce IDTCount, a framework that uses online tracking evidence and completed trajectories to recover, consolidate, and verify identities before computing the final count.
    \item We evaluate identity consistency in video counting across three video instance segmentation datasets, measuring association accuracy, identity switches, and recovery after occlusion.
    \item We also evaluate IDTCount across five unique instance counting benchmarks, spanning long-tailed scenes, dense crowds, and long videos.
\end{itemize}
\section{Related Work}
\label{sec:related}
\subsection{Frame-Level Video Counting}
Frame-level video counting estimates object occupancy at each timestamp, typically exploiting temporal context to stabilize per-frame predictions. Avvenuti et al.~\cite{M_2022_Spatio} use spatiotemporal attention across consecutive frames, LMSTIN~\cite{wang2024lmstin} pairs lightweight feature extraction with multistage temporal inference, and E-MAC~\cite{Cao_2025_Efficient} combines density-embedded masked modeling with optical flow fusion. Alternatively, Shu and Chan~\cite{Shu_2026_CVPR} enforce temporal consistency via a training-time statistical regularizer without inference overhead. Despite these refinements, per-frame occupancy cannot separate returning objects from newly arriving ones, so it cannot determine cumulative instance counts across a video.

\subsection{Tracking and Flow-Based Counting}
To enumerate unique instances across time, cumulative counting methods rely on either trajectory tracking or cross-frame flow estimation. On the tracking side, Wen et al.~\cite{wen2021dronecrowd} jointly optimize detection, tracking, and density estimation in aerial crowds. In contrast, Li et al.~\cite{li2023sceneaware} restrict identity birth and death to scene transition zones to curb matching errors. To avoid explicit trajectory reconstruction, flow-based methods estimate net population changes directly: Fan et al.~\cite{fan2025video} infer inflow and outflow by subtracting shared inter-frame density from global density maps, Liu et al.~\cite{Liu_2024_CVPR} replace trajectory labels with identity-agnostic group contrastive matching, and Huang et al.~\cite{huang2025vicssl} learn cross-frame correspondence through self-supervised motion augmentation. Even so, these methods are largely tailored to single-class pedestrian crowds and rely on either brittle spatial continuity or local flow conservation, both of which fail under prolonged occlusions and arbitrary reentries.

\subsection{Open Vocabulary Video Counting}
Vision-language models enable text-guided detection and counting in static images~\cite{liu2024groundingdino, R_2023_CLIP, J_2023_Zero, M_2024_Language,amininaieni2023countx,amini2024countgd,huang2024pseco}, while promptable video segmentation models such as SAM~2~\cite{ravi2024sam2} and SAM~3~\cite{carion2026sam3} extend prompt-conditioned masks across frames. Building on these capabilities, cumulative video counting must decide whether each observation belongs to a previously counted object or a new instance. CountVid~\cite{amini2025countvid} couples CountGD-Box detections with temporal filtering and SAM~2 ~\cite{ravi2024sam2} mask propagation; yet, it remains vulnerable to missed detections and identity fragmentation across occlusions. SAM3Count~\cite{Owusu_2026_CVPR} attempts to address this and builds a re-identification tracker on top of SAM3's tracker to recover lost tracks due to occlusion. Xu et al.~\cite{xu2026depthguided} count objects given text or visual prompts in RGB-D video, bias the detector's attention with depth, and use depth in track association, followed by temporal voting to remove cross-frame duplicates. In contrast, IDTCount builds on frozen SAM~3 observations to recover persistent identities during tracking and uses evidence from completed trajectories to resolve duplicate identities and suppress spurious ones before computing the final count.

\subsection{Identity Association in MOT and VIS}
Multi-object tracking (MOT) and video instance segmentation (VIS) provide mechanisms for maintaining object identities over time. MOT methods such as GeneralTrack~\cite{Qin_2024_CVPR}, MOTIP~\cite{Gao_2025_CVPR}, and LA-MOTR~\cite{Wang_2025_ICCV} learn associations through relational modeling and trajectory history, while VIS methods, including CTVIS~\cite{Ying_2023_ICCV}, CAVIS~\cite{Lee_2025_CAVIS}, and LOMM~\cite{Lee_2025_LOMM}, leverage temporal context and object memory to preserve correspondence across frames. In addition, complementary approaches address association under challenging observations by incorporating low-confidence detections in ByteTrack~\cite{zhang2022bytetrack} or extending tracking and segmentation across open categories in OVTrack~\cite{li2023ovtrack}, OV2Seg~\cite{wang2023towards}, and MASA~\cite{li2024masa}. Beyond associating frame-level observations, reliable video predictions require assessing accumulated tracks and resolving redundancy. Brissman et al.~\cite{brissman2023recurrent} score track confidence from temporal evidence to filter unreliable predictions, while Propose-Reduce~\cite{lin2021propose} suppresses redundant instance sequences belonging to the same object. These mechanisms are critical for cumulative counting, where fragmented trajectories inflate counts and false associations merge distinct instances. IDTCount unifies online association with trajectory-level refinement in a persistent counting pipeline, pairing a learned association module that reactivates dormant identities or initiates new ones with learned verification and geometric duplicate consolidation to determine the final count.

\section{Methodology}
\label{sec:methodology}

\begin{figure*}[t]
\centering
\includegraphics[width=\linewidth]{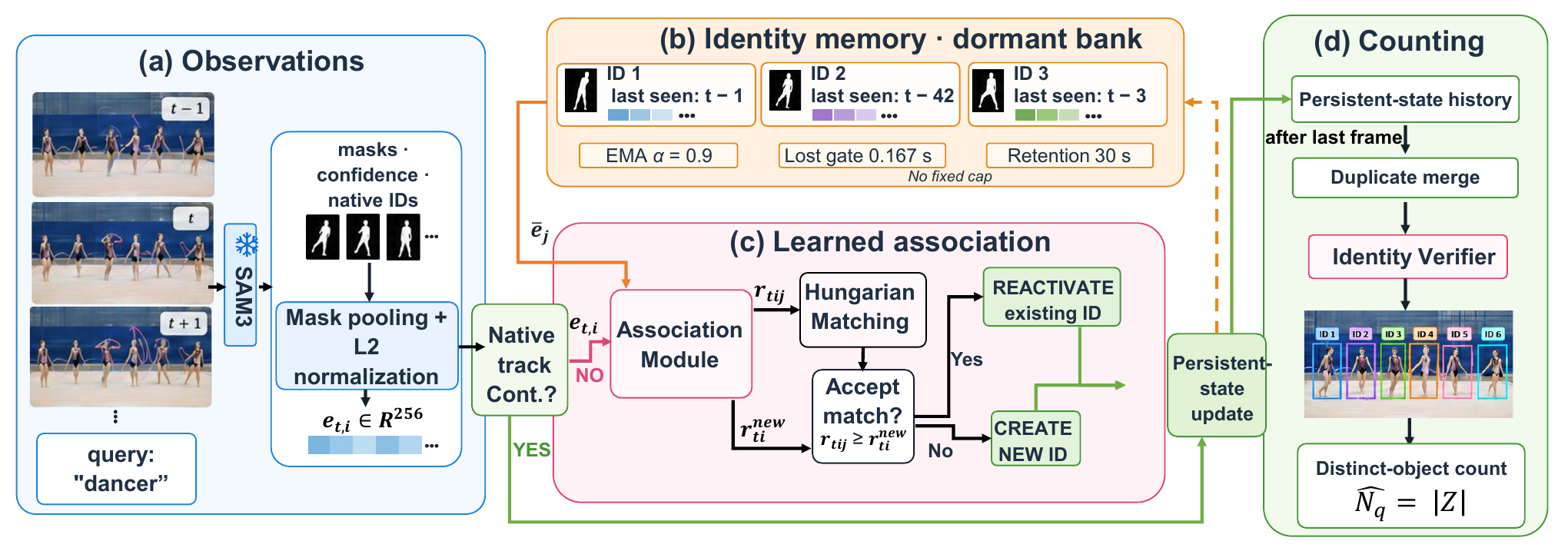}
\caption{
IDTCount's architecture. Given a video and a text prompt, the frozen baseline produces per-object observations with masks, scores, mask-pooled features, and native track states. For each observation that needs a decision, the association module scores every eligible dormant identity, and one-to-one matching within the frame either reactivates a dormant identity or creates a new one. After the last frame, the duplicate merge and a verifier yield the final count $\widehat N_q$.
}
\label{fig:architecture}
\end{figure*}

\subsection{Problem Formulation}
\label{sec:problem_formulation}

Let $V=(I_1,\ldots,I_T)$ denote a video, and let $q$ be an arbitrary natural-language query. The goal is to estimate the number of distinct physical instances matching $q$ that appear at least once in the video:
\begin{equation}
N_q =
\left|
\left\{
o \; \middle| \;
\exists\, t \in \{1,\ldots,T\},
\;
o \text{ is visible in } I_t
\text{ and }
o \models q
\right\}
\right|.
\end{equation}

At each frame $t$, we adopt SAM3 as our baseline~\cite{carion2026sam3} to produce a set of text-conditioned observations:
\begin{equation}
D_t(q) = \left\{ d_{t,i} \right\}_{i=1}^{n_t}.
\end{equation}
Each observation contains a mask, a confidence score, a native track ID, and an internal visual representation. We assign a persistent identity $z_{t,i}$ to each accepted observation, which is the one with sufficient confidence and a mask large enough to pool a feature. The predicted final count then becomes the size of the identity set after a duplicate merge, which joins identities that describe the same object:
\begin{equation}
\widehat{N}_q = |Z|,
\quad
Z = \operatorname{merge}\!\left(
\bigcup_{t=1}^{T}
\left\{
z_{t,i}
\;\middle|\;
d_{t,i} \text{ accepted}
\right\}
\right).
\end{equation}

\subsection{Architecture}
\label{sec:architecture}
IDTCount is built on a frozen SAM3 baseline that provides text-conditioned detections, masks, visual features, and native tracks. Its identity layer maintains the persistent identities it counts. The four panels in Figure~\ref{fig:architecture} show observation extraction, dormant memory, learned association, and persistent counting. The compact association block contains jointly trained pairwise and new-identity scorers. An observation that continues a native track retains the identity of that track. Any other observation goes to the association module, which decides whether to reactivate a dormant identity or take a new one. 

\subsubsection{Appearance Features}
\label{sec:appearance}
For an accepted observation, $d_{t, i}$, let $a_{t, i} \in \mathbb{R}^{256}$ denote the backbone feature pooled over its predicted mask, without any learned projection, and let $e_{t,i}=a_{t,i}/\|a_{t,i}\|_2$. We use the baseline's own features already computed for every frame. For an identity bank entry $j$ with prototype $\bar{e}_j$, appearance similarity is
\begin{equation}
c_{ij}
=
e_{t,i}^{\top}\bar{e}_j.
\end{equation}

\subsubsection{Association Module}
\label{sec:head2}

The association module compares reactivation against creating a new identity with two jointly trained heads. For each observation--candidate pair $(i,j)$, a pairwise head receives a 32-dimensional vector
\begin{equation}
x_{tij}=\left[v_{tij};m_{tij}\right]\in\mathbb{R}^{32},
\end{equation}
where $v_{tij}\in\mathbb{R}^{16}$ contains association cues (appearance similarity, geometric alignment, temporal gap, motion consistency, detection and tracking confidence, and competition with other candidates), and $m_{tij}\in\mathbb{R}^{16}$ marks which cues are available and outputs a reactivation logit $r_{tij}$. An unavailable cue is set to zero together with its indicator, so a zero is never mistaken for a measurement. A second, \textsc{New} head outputs a \textsc{New} logit $r_{ti}^{\mathrm{new}}$ from a summary of the candidate set. This set includes detection confidence, candidate count, the two largest pair logits, and the smallest motion residual. The \textsc{New} score sees the candidate evidence, so it rises when no candidate resembles the observation. Both are trained jointly with cross-entropy over the dynamic candidate set $\mathcal{C}_{t,i}$:
\begin{equation}
\mathcal{L}_{ti}=-\log
\frac{\exp\left(r_{ti,y_{ti}}\right)}{\exp\left(r_{ti}^{\mathrm{new}}\right)+\sum_{j\in\mathcal{C}_{t,i}}\exp\left(r_{tij}\right)},
\end{equation}
where $y_{ti}\in\mathcal{C}_{t,i}\cup\{\mathrm{new}\}$ is the target decision and $r_{ti,y_{ti}}=r_{ti}^{\mathrm{new}}$ when $y_{ti}=\mathrm{new}$. The module receives no category name or identifier.

\paragraph{Supervision:} The targets $y_{ti}$ come from ground-truth identity tracks that label decisions and never serve as inputs. We replay a training video through the frozen baseline and identity bank to trigger every decision that requires the module. The target is the candidate that holds the observation's ground-truth identity (\textsc{Match}), or \textsc{New} if no candidate does. Skip a decision if its matching identity is not in the bank.

\subsection{Dormant Identity Bank}
\label{sec:identity_bank}
The identity bank maintains local track states alongside their mappings to persistent counting identities. Each entry stores an appearance prototype $\bar e_j$, recent box history, the last observation time, and available confidence and visibility statistics. When observing the corresponding local track, the prototype updates via a normalized exponential moving average,
\begin{equation}
 \bar e_j\leftarrow\operatorname{norm}\left(\alpha\bar e_j+(1-\alpha)e_{ti}\right),\qquad\alpha=0.9.
\end{equation}
While appearance states belong to local tracks, the persistent mapping allows several successive local tracks to represent a single counted identity.

A missing local track becomes eligible for dormant recovery after a configured lost track gate of nominally $0.167$ s and at least one processed frame. Entries carry a retention window of $30$ s. For any fresh observation, the candidate set $\mathcal C_{ti}$ consists of eligible unexpired dormant entries mapped to a persistent identity. Our configuration applies no fixed candidate cap or spatial filter, allowing appearance cues to drive recovery even after substantial displacement. Expiring an entry simply removes a recovery candidate rather than deleting the identity from the historical count.

\subsubsection{Joint Assignment and Persistent State Update}
\label{sec:joint_assignment}
Multiple observations in a single frame can resemble the same dormant identity, yet a dormant identity can return as only one physical object. We adopt Hungarian matching over the reactivation logits $R_t=[r_{tij}]$ to pair each observation with at most one candidate and each candidate with at most one observation. Ineligible pairs with $j \notin \mathcal{C}_{t,i}$ are masked out. A matched pair is committed as a reactivation only if $r_{tij}\geq r_{ti}^{\mathrm{new}}$, so the learned \textsc{New} logit acts as the acceptance threshold; otherwise, the observation creates a new identity, as do observations that lose the matching or have no candidates. Native-track continuations skip this step, and all paths update the persistent state.

\subsection{Post-Tracking Merging and Verification}
\label{sec:post_tracking}
Association can still produce spurious tracks or split one physical object across multiple identities. After the final frame, we refine the identity set $Z_{\mathrm{raw}}$ in two steps before computing the count. We first consolidate $Z_{\mathrm{raw}}$ by merging trajectories that exhibit sustained spatial overlap in shared frames or satisfy short-gap handover criteria, producing the consolidated identity set $Z$. We then apply a trajectory-level verifier to filter out unreliable identities. For each identity $k \in Z$, a gradient-boosted classifier estimates the probability $p_k$ that the track is spurious or a duplicate:
\begin{equation}
    p_k = f_{\psi}(\phi_k),
\end{equation}
where $\phi_k \in \mathbb{R}^{18}$ summarizes the track's temporal support, spatial overlap with neighboring identities, and creation association scores. 

We train the classifier on predicted identities from OVIS-valid~\cite{qi2022occluded}, DanceTrack-train~\cite{sun2022dancetrack}, and MOT17-train~\cite{dendorfer2021motchallenge}, with ground-truth tracks providing valid, duplicate, and spurious labels.  Video-disjoint selection data determine the feature configuration and decision threshold of $0.5$, which are fixed at inference.
Verification produces the retained set $Z' \subseteq Z$, and the final predicted count is $\widehat{N}_q = |Z'|$. We provide feature definitions, training details, and threshold selection in the supplementary material.

\subsection{Inference}
\label{sec:inference}
For each video and text-query pair, we run the baseline with the query and pass accepted observations through the identity layer. After the last frame, the duplicate merge and verifier produce the set $Z'$ of persistent identities that remain, and the final prediction is $\widehat{N}_q = |Z'|$. A video that is too long or too crowded for one tracking session in memory is processed in overlapping sessions. Identities continue across a session boundary by one-to-one matching of tracks on the overlapping frames, and all other observations of a new session pass through the association module as above. SAM3's tracker memory otherwise grows with every processed frame, so we run the tracker with eviction of frames that have left its attention window. This is a setting of the frozen tracker, not a learned component, and it is also used when generating the decisions on which the association module is trained, so training and inference see the same observations.


\section{Experiments}
\label{sec:experiments}
\begin{table*}[!h]
\centering
\caption{
Identity association on the identity benchmarks. Best results shown in bold.}
\label{tab:identity_main}
\small
\setlength{\tabcolsep}{3.2pt}
\renewcommand{\arraystretch}{1.08}
\begin{tabular}{@{}llcccccc@{}}
\toprule
Dataset &
Method &
AssA (\%) $\uparrow$ &
HOTA (\%) $\uparrow$ &
DetA (\%) $\uparrow$ &
IDF1 (\%) $\uparrow$ &
ID Sw. $\downarrow$ &
Long Rec. (\%) $\uparrow$ \\
\midrule
\multirow{6}{*}{OVIS-test} & CountVid~\cite{amini2025countvid} (AAAI '26) & 81.21 & 65.79 & 53.50 & 70.92 & 484 & 76.64 \\
 & ByteTrack~\cite{zhang2022bytetrack} (ECCV '22) & 63.98 & 67.23 & 70.92 & 70.01 & 1300 & 14.46 \\
 & MASA~\cite{li2024masa} (CVPR '24) & 63.04 & 67.26 & 72.10 & 69.62 & 2196 & 45.19 \\
 & SAM3~\cite{carion2026sam3} (ICLR '26) & 86.54 & 78.31 & 71.02 & 84.93 & 308 & \textbf{82.52} \\
 & SAM3Count~\cite{Owusu_2026_CVPR} & 83.02 & 75.90 & 69.55 & 83.94 & 199 & 80.77 \\
 & IDTCount (ours) & \textbf{87.23} & \textbf{80.24} & \textbf{73.97} & \textbf{87.48} & \textbf{169} & 80.95 \\
\midrule
\multirow{6}{*}{YT-VIS} & CountVid~\cite{amini2025countvid} (AAAI '26) & 81.91 & 57.54 & 40.76 & 58.60 & 312 & 77.11 \\
 & ByteTrack~\cite{zhang2022bytetrack} (ECCV '22) & 68.19 & 63.30 & 58.91 & 64.01 & 1432 & 7.41 \\
 & MASA~\cite{li2024masa} (CVPR '24) & 72.80 & 65.48 & 59.07 & 66.43 & 1939 & 57.55 \\
 & SAM3~\cite{carion2026sam3} (ICLR '26) & 88.51 & 70.86 & 56.84 & 74.51 & 249 & 76.67 \\
 & SAM3Count~\cite{Owusu_2026_CVPR} & 81.12 & 68.44 & 57.87 & 75.37 & 195 & 66.67 \\
 & IDTCount (ours) & \textbf{89.14} & \textbf{72.53} & \textbf{59.13} & \textbf{76.98} & \textbf{128} & \textbf{86.21} \\
\midrule
\multirow{6}{*}{LV-VIS} & CountVid~\cite{amini2025countvid} (AAAI '26) & 89.43 & 49.81 & 28.09 & 44.98 & 97 & 81.67 \\
 & ByteTrack~\cite{zhang2022bytetrack} (ECCV '22) & 86.85 & 71.52 & 59.07 & 73.20 & 899 & 29.82 \\
 & MASA~\cite{li2024masa} (CVPR '24) & 89.32 & 73.08 & 59.98 & 75.56 & 669 & 87.13 \\
 & SAM3~\cite{carion2026sam3} (ICLR '26) & 94.60 & 77.11 & 63.05 & 79.48 & 90 & 86.96 \\
 & SAM3Count~\cite{Owusu_2026_CVPR} & 82.55 & 67.11 & 54.71 & 72.68 & \textbf{44} & 91.18 \\
 & IDTCount (ours) & \textbf{94.62} & \textbf{77.62} & \textbf{63.87} & \textbf{80.09} & 68 & \textbf{92.86} \\
\bottomrule
\end{tabular}
\end{table*}
IDTCount is a complete counting system combining a frozen SAM3 detector and tracker~\cite{carion2026sam3} with our identity layer, which recovers object identities across occlusions and reentries. Because this identity layer can also be attached to other trackers, we evaluate three aspects of our approach. First, we evaluate the full system on unique instance counting against CountVid~\cite{amini2025countvid}, SAM3Count~\cite{Owusu_2026_CVPR}, and SAM3 across five counting benchmarks spanning dense scenes and long videos. Second, we assess identity consistency against those same baselines as well as ByteTrack~\cite{zhang2022bytetrack} and MASA~\cite{li2024masa} multi-object trackers. These run on SAM3 detections across three video instance segmentation benchmarks. Third, we evaluate the identity layer as a modular extension on top of ByteTrack and MASA with the SAM3 detector. We additionally evaluate detector choice and transfer to CountVid on Crystals, followed by component ablations.

\subsection{Datasets and Metrics}
\label{sec:datasets}
\paragraph{Datasets:} We evaluate on four benchmarks from CountVid's VideoCount dataset~\cite{amini2025countvid}. TAO-Count tests low-count scenes with substantial motion; MOT20-Count covers heavily crowded pedestrian scenes; Penguins contains three videos of look-alike birds; and Crystals contains seven scientific videos of visually similar crystals, which we also use to examine transfer across detectors. We additionally construct BURST-Count from the BURST test split~\cite{Athar_2023_BURST}, following TAO-Count's construction ~\cite{amini2025countvid}. It evaluates counting across a broad, long-tailed vocabulary with motion, occlusions, and re-entries. AnimalTrack-Count comprises the 25 AnimalTrack test videos~\cite{zhang2023animaltrack} in which every animal of the queried species is labeled, testing dense groups of visually similar animals. We then assess the persistent identity tracking quality of IDTCount on three video instance segmentation (VIS) benchmarks, using the dataset category names as text prompts. These are OVIS-test~\cite{qi2022occluded}, with frequent occlusion and object overlap; YouTube-VIS (YT-VIS)~\cite{Yang_2019_YTVIS}, with broad scene diversity; and LV-VIS~\cite{wang2023towards}, which tests open-vocabulary association. A video-level count depends on the persistent identity tracks behind it, so these benchmarks let us measure association quality rather than relying only on the final count metrics. 
\begin{figure*}[!h]
\centering
\includegraphics[width=\textwidth]{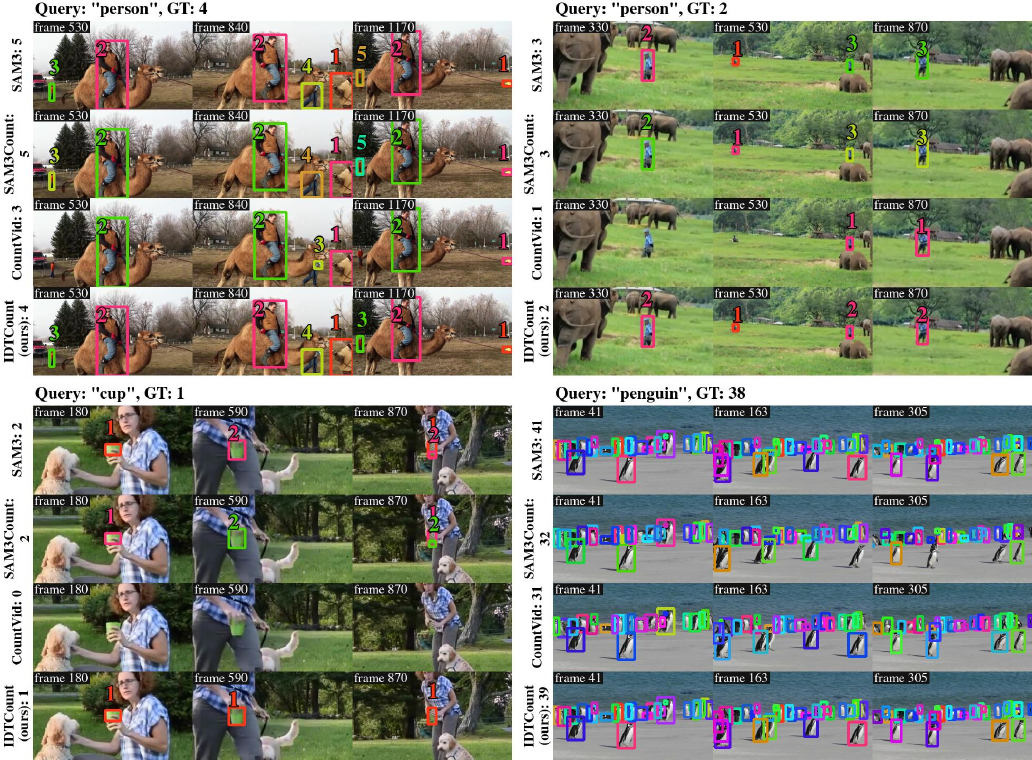}
\caption{
Qualitative comparison of SAM3~\cite{carion2026sam3}, SAM3Count~\cite{Owusu_2026_CVPR}, CountVid~\cite{amini2025countvid}, and IDTCount on three BURST-Count videos and one dense AnimalTrack-Count video (bottom right).}
\label{fig:qualitative_comparison}
\end{figure*}
\paragraph{Metrics:} For counting, we report the video-level MAE and RMSE defined by CountVid~\cite{amini2025countvid}, $\mathrm{MAE} = \frac{1}{M}\sum_{m=1}^{M} |\widehat{N}_m - N_m|$ and $\mathrm{RMSE} = (\frac{1}{M}\sum_{m=1}^{M} (\widehat{N}_m - N_m)^2)^{1/2}$ over $M$ video-query pairs, where the ground truth $N_m$ is the number of unique objects, not the number of detections. We also report the fractions of over- and undercounted queries. For identity quality, we report AssA, HOTA, DetA~\cite{luiten2021hota}, and IDF1~\cite{Ristani_2016_IDF1}, the number of ID switches (changes of the predicted identity matched to a ground-truth object), and long-gap recovery (the fraction of objects that, after being missed for at least 1\,s, return with the same identity as before).

\begin{table*}[t]
\centering
\caption{Unique-instance counting across five benchmarks. Lower MAE and RMSE are better. Dashes indicate unavailable results.}
\label{tab:counting_results}
\small
\setlength{\tabcolsep}{4pt}
\renewcommand{\arraystretch}{1.08}
\resizebox{\textwidth}{!}{%
\begin{tabular}{@{}lcccccccccc@{}}
\toprule
& \multicolumn{2}{c}{BURST-Count}
& \multicolumn{2}{c}{TAO-Count}
& \multicolumn{2}{c}{Penguins}
& \multicolumn{2}{c}{MOT20-Count}
& \multicolumn{2}{c}{AnimalTrack-Count} \\
\cmidrule(lr){2-3}\cmidrule(lr){4-5}\cmidrule(lr){6-7}
\cmidrule(lr){8-9}\cmidrule(lr){10-11}
Method
& MAE $\downarrow$ & RMSE $\downarrow$
& MAE $\downarrow$ & RMSE $\downarrow$
& MAE $\downarrow$ & RMSE $\downarrow$
& MAE $\downarrow$ & RMSE $\downarrow$
& MAE $\downarrow$ & RMSE $\downarrow$ \\
\midrule
CountVid~\cite{amini2025countvid}
& 1.433 & 4.750 & 2.595 & 5.894 & 4.0 & 5.3 & 50.0 & 61.3 & 8.7 & 14.5 \\
SAM3~\cite{carion2026sam3}
& 0.654 & 1.792 & 0.653 & 1.606 & 2.7 & 3.4 & -- & -- & -- & -- \\
SAM3Count~\cite{Owusu_2026_CVPR}
& 0.594 & 1.525 & 0.78 & 1.63 & 2.7 & 3.2 & -- & -- & -- & -- \\
IDTCount (ours)
& \textbf{0.537} & \textbf{1.301}
& \textbf{0.565} & \textbf{1.314}
& \textbf{2.0} & \textbf{2.9}
& \textbf{11.0} & \textbf{12.9}
& \textbf{7.6} & \textbf{11.2} \\
\bottomrule
\end{tabular}%
}
\end{table*}

\subsection{Implementation Details}
\label{sec:implementation}
The association module is trained on identity decisions replayed from the training splits of OVIS~\cite{qi2022occluded}, YouTube-VIS~\cite{Yang_2019_YTVIS}, and LV-VIS~\cite{wang2023towards}, together with scenarios S01 and S04 of AIC22 CityFlowV2~\cite{Tang_2019_CVPR,Naphade_2022_CVPR}. We balance reactivation and new identity decisions equally and optimize with AdamW for up to 100 epochs. Using held-out validation data, we select the checkpoint that achieves the highest worst-source reactivation accuracy while staying below a ceiling on false reactivations. We do not use any of the evaluation splits during training or model selection. At inference time, observations with confidence below $\tau$  are discarded, videos are sampled at each benchmark's frame rate, and the tracker evicts stale frames from its memory. Further training and selection details are in the supplementary material.

\subsection{Quantitative Results}
\label{sec:identity_results}
\label{sec:counting_results}

Table~\ref{tab:counting_results} compares unique-instance counting across five benchmarks. On BURST-Count, TAO-Count, and Penguins, IDTCount achieves the lowest MAE and RMSE, outperforming CountVid, SAM3, and SAM3Count. These results demonstrate improved final count accuracy across both diverse scenes and look-alike objects. On the dense-scene benchmarks MOT20-Count and AnimalTrack-Count in Table~\ref{tab:counting_results}, IDTCount achieves lower MAE and RMSE than CountVid, with particularly substantial gains on MOT20-Count. As released, SAM3 and SAM3Count exceed the memory capacity of an 80\,GB H100 on dense videos in MOT20-Count (52--398 processed frames and 80--1,203 ground-truth objects per video) and AnimalTrack-Count (196--2,126 processed frames and 8--134 ground-truth objects per video), so their results are not reported on these benchmarks. Table~\ref{tab:identity_main} evaluates identity consistency on OVIS.test, YT-VIS, and LV-VIS. IDTCount achieves the highest AssA, HOTA, DetA, and IDF1 across all three benchmarks. ByteTrack and MASA, operating on the same SAM3 detections, achieve lower association accuracy, showing that detection quality alone does not ensure consistent identities. IDTCount also records the fewest identity switches on OVIS.test and YT-VIS, while SAM3Count records fewer on LV-VIS. For recovery after long occlusions, IDTCount performs best on YT-VIS and LV-VIS, whereas SAM3 leads on OVIS-test.

\begin{table}[!h]
\centering
\caption{Counting on Crystals dataset. CVD: CountVid; CGD-BX: CountGD-Box. IR- Internal Exemplars.}
\label{tab:counting_crystals}
\small
\setlength{\tabcolsep}{3pt}
\renewcommand{\arraystretch}{1.08}
\begin{tabular}{@{}p{0.40\columnwidth}cccc@{}}
\toprule
& \multicolumn{2}{c}{Text only}
& \multicolumn{2}{c}{\shortstack{Text + IR}} \\
\cmidrule(lr){2-3}\cmidrule(lr){4-5}
Method & MAE $\downarrow$ & RMSE $\downarrow$
& MAE $\downarrow$ & RMSE $\downarrow$ \\
\midrule
SAM3 & 74.71 & 89.45 & 74.71 & 89.45 \\
SAM3 + IDTCount & 74.71 & 89.45 & 74.71 & 89.45 \\
CountVid~\cite{amini2025countvid} & 69.00 & 86.01 & 27.86 & 45.04 \\
CVD + IDTCount & 64.86 & 83.23 & 23.86 & 36.27 \\
CGD-BX + SAM3 + IDT & \textbf{64.71} & \textbf{83.09} & \textbf{17.29} & \textbf{32.00} \\
\bottomrule
\end{tabular}
\end{table}

\begin{table*}[!h]
\centering
\caption{IDTCount's association layer in other trackers (MAE / RMSE) on SAM3 detections; best in bold.}
\label{tab:plugin_counting}
\small
\setlength{\tabcolsep}{3.4pt}
\begin{tabular}{@{}lcccccccc@{}}
\toprule
& \multicolumn{2}{c}{BURST-Count} & \multicolumn{2}{c}{TAO-Count} & \multicolumn{2}{c}{OVIS-test} & \multicolumn{2}{c}{LV-VIS} \\
\cmidrule(lr){2-3}\cmidrule(lr){4-5}\cmidrule(lr){6-7}\cmidrule(lr){8-9}
Method & MAE $\downarrow$ & RMSE $\downarrow$ & MAE $\downarrow$ & RMSE $\downarrow$ & HOTA (\%) $\uparrow$ & IDF1 (\%) $\uparrow$ & HOTA (\%) $\uparrow$ & IDF1 (\%) $\uparrow$ \\
\midrule
ByteTrack & 4.106 & 8.243 & 3.892 & 7.797 & 67.23 & 70.01 & 71.52 & 73.20 \\
ByteTrack + IDTCount & \textbf{2.640} & \textbf{4.816} & \textbf{2.552} & \textbf{4.696} & \textbf{68.62} & \textbf{73.06} & \textbf{71.81} & \textbf{73.85} \\
\midrule
MASA & 2.650 & 6.386 & 2.482 & 5.659 & 67.26 & 69.62 & 73.08 & 75.56 \\
MASA + IDTCount & \textbf{2.338} & \textbf{5.112} & \textbf{2.202} & \textbf{4.717} & \textbf{67.38} & \textbf{70.40} & \textbf{73.19} & \textbf{75.77} \\
\bottomrule
\end{tabular}
\end{table*}

\paragraph{Crystals:} Table~\ref{tab:counting_crystals} examines detector choice and identity-layer transfer on scientific imagery. We evaluate text-only prompting and text augmented with internal exemplars (up to three highest-confidence valid detection boxes selected across the same video's text-only detector pass), without ground-truth annotations. The SAM3 detector produces no detections on these seven videos, so both the native SAM3 control and its IDTCount extension return zero counts. We supply CountGD-Box detections to SAM3 and apply IDTCount to the resulting tracks. This configuration achieves the lowest MAE and RMSE among the configurations shown, with substantially lower errors when internal exemplars are provided. We also replace CountVid's temporal filter with IDTCount's identity handling while retaining its CountGD-Box detector and SAM2 tracker. This replacement reduces both MAE and RMSE relative to native CountVid under both prompt settings.

\begin{figure}[!h]
\centering
\includegraphics[width=\columnwidth]{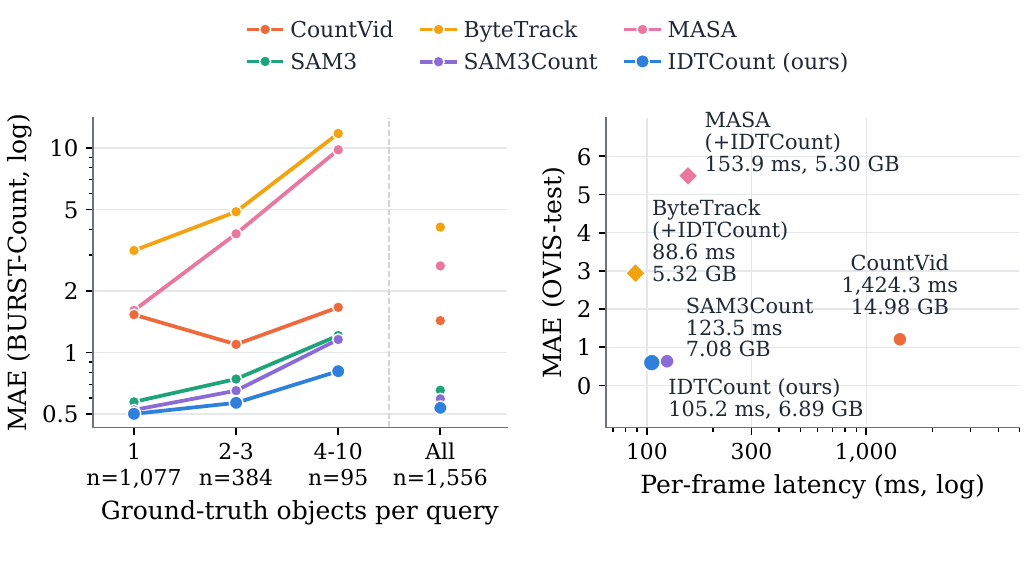}
\caption{Left: counting MAE on BURST-Count grouped by ground truth object count (log scale), where All denotes the full evaluation set. Right: accuracy and efficiency trade-off comparing counting MAE against per-frame latency on OVIS-test.}
\label{fig:analysis_pair}
\end{figure}

\paragraph{Integrating IDTCount as an Identity Layer:} Table~\ref{tab:plugin_counting} evaluates IDTCount's identity layer performance when paired with alternative trackers in place of SAM3's native tracker. Using the same SAM3 detections as input, we run ByteTrack~\cite{zhang2022bytetrack} and MASA~\cite{li2024masa} both with and without our identity layer. ``+IDTCount'' replaces the tracker's identity handling. Figure~\ref{fig:analysis_pair} uses the same pairing, with each method pair (e.g., ByteTrack (+IDTCount)) showing the tracker and the tracker with this layer. IDTCount reduces both trackers’ counting errors on BURST-Count and TAO-Count, showing gains beyond SAM3.

\subsection{Qualitative Analysis}
\label{sec:qualitative}
Figure~\ref{fig:qualitative_comparison} compares SAM3~\cite{carion2026sam3}, SAM3Count~\cite{Owusu_2026_CVPR}, CountVid~\cite{amini2025countvid}, and IDTCount across four representative videos illustrating scene reentry, small distant objects, viewpoint changes, and dense crowds. In the first sequence, a small person standing near a truck exits the frame and later returns. Both SAM3 and SAM3Count assign a new identity upon reentry, double-counting the same individual, while CountVid fails to detect the person at all. IDTCount recovers the dormant bank entry and reconnects the person to the original identity. In the second sequence, a walking man passes behind elephants before reappearing. Both SAM3 and SAM3Count fragment his trajectory into a new identity after the occlusion, while CountVid misses a seated person throughout the clip. IDTCount preserves stable identities for both individuals across the entire sequence. In the third sequence, a woman walks a dog while carrying a single cup. After an abrupt viewpoint change, SAM3 and SAM3Count assign the cup to a new identity, while CountVid misses it even when unoccluded. IDTCount maintains a single persistent identity for the cup throughout. The fourth sequence depicts a dense colony of overlapping penguins moving through the scene. SAM3 slightly overcounts due to track fragmentation, whereas CountVid and SAM3Count undercount. IDTCount tracks the colony consistently and stays closest to the ground-truth count. Additional videos are in the supplementary material.

\subsection{Ablations}
\label{sec:head2_necessity}
Table~\ref{tab:head2_burst_ablation} adds IDTCount's components one at a time to the frozen SAM3 tracker and scores identity quality on BURST-Count. The association module lowers ID switches from 341 to 327, and the geometric merge changes them by one (326). The verification stage gives a large reduction, to 186 ID switches, and raises AssA.\label{sec:verifier}
\begin{table}[!h]
\centering
\caption{Cumulative component ablation on BURST-Count}
\label{tab:head2_burst_ablation}
\small
\setlength{\tabcolsep}{6pt}
\renewcommand{\arraystretch}{1.08}
\begin{tabular}{@{}lcc@{}}
\toprule
Variant & AssA (\%) $\uparrow$ & ID Sw. $\downarrow$ \\
\midrule
SAM3 & 29.68 & 341 \\
+ association module & 29.67 & 327 \\
+ merge & 29.67 & 326 \\
+ verifier & \textbf{29.88} & \textbf{186} \\
\bottomrule
\end{tabular}
\end{table}

\subsection{Count Error Analysis}
\label{sec:analysis}
Figure~\ref{fig:analysis_pair} (left) partitions BURST-Count by ground-truth object count, with the overall error shown on the right. ByteTrack and MASA degrade rapidly as object density increases, whereas CountVid remains flat at a high error level. IDTCount achieves the lowest error across every bin and overall, outperforming the SAM3 tracker.

\subsection{Efficiency, Runtime, and Memory}
\label{sec:efficiency}
IDTCount trains 13,026 parameters in its association module, plus a frozen identity verifier that runs on CPU (200 depth-3 trees over 18 features). The SAM3 backbone is not fine-tuned. CountVid fine-tunes the MLP box heads of CountGD-Box~\cite{amini2025countvid}; based on its released checkpoint, these are two three-layer MLPs with 265,224 parameters in total, about 20 times more than IDTCount's association module. We measure per-frame latency as the median time per frame of the tracking loop, averaged over the 224 OVIS-test queries; CountVid, whose pipeline is not streaming, is measured by its inference time per frame. IDTCount is much faster than CountVid and uses less peak GPU memory. Its reported latency covers the identity bank, the association module, and the duplicate merge. Post-video verification runs separately on CPU and takes 4.42\,s across all 224 OVIS-test queries, including trajectory-feature computation and classifier inference. Figure~\ref{fig:analysis_pair} (right) compares only counting systems, plotting per-frame latency against OVIS-test count error.
\section{Conclusion}
\label{sec:conclusion}

We introduced IDTCount, an identity-driven temporal counting framework that maintains persistent object identities across frames rather than aggregating per-frame detections. Across three identity association benchmarks and five unique instance counting benchmarks, IDTCount substantially improves both tracking consistency and final count accuracy over prior open vocabulary counters while adding minimal computational overhead. These results confirm that framing cumulative video counting as a persistent identity assignment problem effectively prevents overcounting across occlusions and reentries. A remaining limitation is reliance on observations from a frozen backbone, as IDTCount can link observed instances but cannot recover objects missed throughout a video. Enhancing backbone detection coverage in dense crowds while preserving open vocabulary generalization is therefore a promising direction for future work.
\section*{LLM Usage Statement}

We used two AI tools. \emph{Claude} (Anthropic), a large language model, helped us plan the paper's structure and draft it, which the authors then revised. \emph{Grammarly} (its AI writing assistance) was used to correct grammar. The authors take full responsibility for all content.

{
    \small
    \bibliographystyle{ieeenat_fullname}
    \bibliography{main}
}

\clearpage
\setcounter{page}{1}
\maketitlesupplementary

\setcounter{section}{0}
\renewcommand{\thesection}{\Alph{section}}
\renewcommand{\theHsection}{supp.\Alph{section}} 

This supplementary material provides further details on IDTCount and its evaluation. Sec.~\ref{sec:supp_method} describes the persistent identity state, association module, duplicate merge, and trajectory-level verifier in full. Sec.~\ref{sec:supp_training} gives the training, checkpoint-selection, and evaluation protocols. Sec.~\ref{sec:supp_data} describes how we build BURST-Count and the other counting benchmarks. Sec.~\ref{sec:supp_results} reports results for dense scenes and long videos, along with efficiency measurements, and Sec.~\ref{sec:supp_qualitative} shows further qualitative examples.

\section{Further Method Details}
\label{sec:supp_method}

\subsection{Persistent Identity State}

\begin{figure*}[t]
\centering
\includegraphics[width=\textwidth]{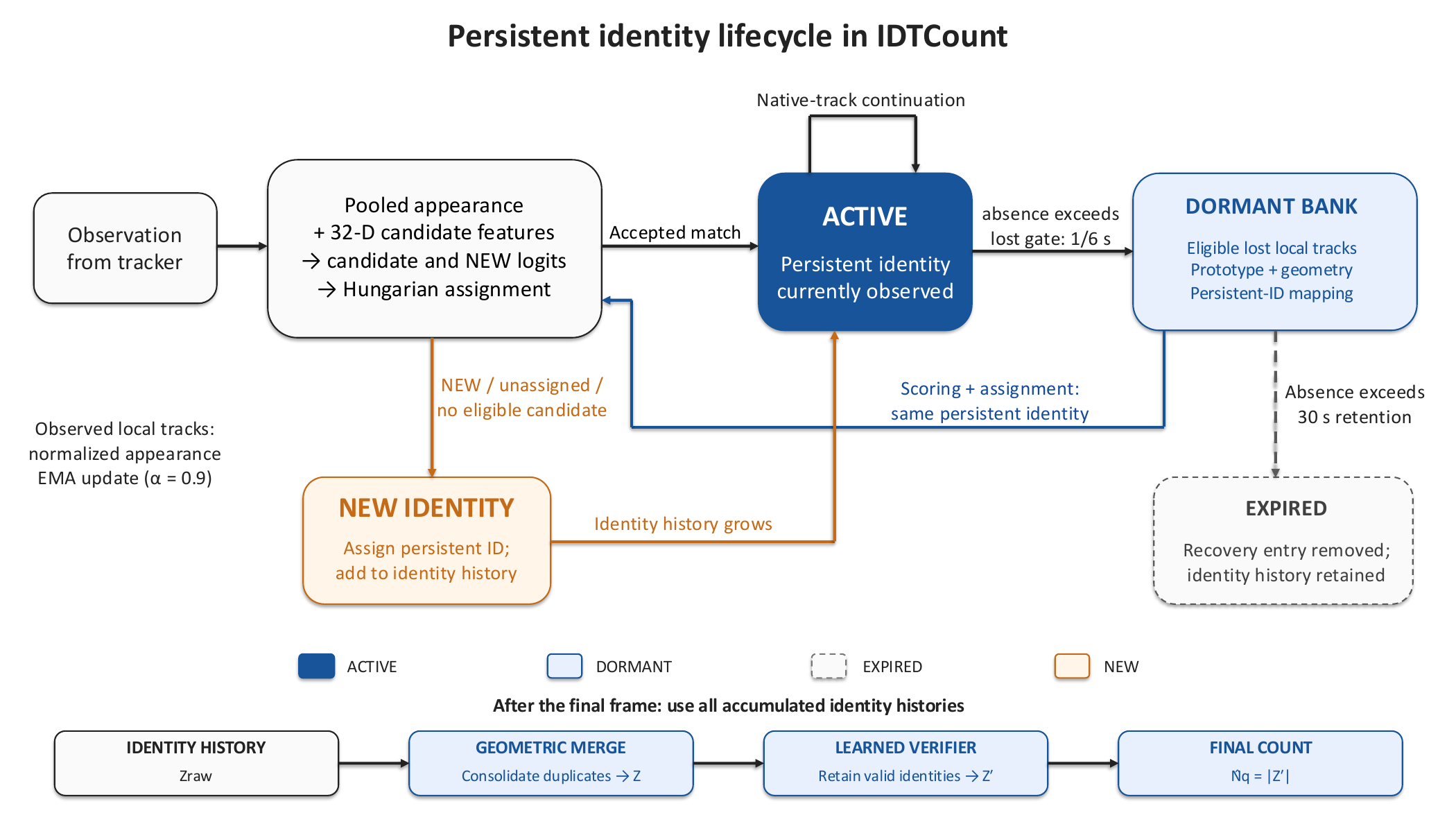}
\caption{Persistent-identity lifecycle in IDTCount, from online association and dormant-identity recovery to post-video merging and verification for the final count.}
\label{fig:supp_identity_lifecycle}
\end{figure*}

Figure~\ref{fig:supp_identity_lifecycle} summarizes the online identity lifecycle and the post-video stages that determine the final count.

At time $t$, IDTCount maintains the set of active identities $\mathcal{A}_t$ and the identity bank $\mathcal{B}_t$, which holds one entry per lost native track. Each persistent identity maintains its identifier, most recent mask and box, first and last seen times, confidence, recent occlusion fraction, and the box history used for constant-velocity prediction. A native track enters the association bank once it receives no accepted observation for the lost-track gate of 0.167\,s, corresponding to $\max(1,\operatorname{round}(0.167r))$ frames at a processing rate of $r$. Each bank entry tracks an appearance prototype alongside a retention clock. After 30\,s of absence, the entry is removed from the bank and becomes ineligible for reactivation.

For each observation requiring a decision, IDTCount computes an association vector against every eligible bank entry. The association module scores all resulting pairs, and a frame-wide Hungarian assignment over the pair logits resolves competition between observations. Each matched pair is then compared against that observation's \textsc{New} logit. Decisions are committed immediately without a provisional confirmation stage or a post hoc probability threshold.

\paragraph{Retention window:} We retain dormant entries for 30\,s to support recovery across extended absences while bounding candidate history. After 30\,s of absence, we remove an entry from the association bank, while its persistent identity remains in the counting history. Table~\ref{tab:supp_retention} compares alternative retention windows and lost-track gates.

\subsection{Appearance Features}
\label{sec:supp_appearance}

Let $F_t \in \mathbb{R}^{H' \times W' \times 256}$ be the frozen backbone feature map, and let $M_i^t$ be the predicted mask for observation $i$, resized to the resolution of $F_t$. Mask pooling gives
\begin{equation}
  a_i^t = \frac{\sum_{u,v} M_i^t(u,v)\, F_t(u,v)}{\sum_{u,v} M_i^t(u,v) + \epsilon} \in \mathbb{R}^{256},
\end{equation}
which we L2-normalize, so appearance similarity is the cosine similarity. In the SAM3-based implementation, mask pooling reuses the backbone feature map already computed for tracking and requires no additional network forward pass.

\subsection{Association Module}
\label{sec:supp_head2}

For each observation--candidate pair $(i,k)$, the association module receives 16 cue values and their binary availability indicators, giving $x_{ik}\in\mathbb{R}^{32}$. Unavailable cues have value zero and availability zero; observed cues have availability one. Table~\ref{tab:supp_features} defines the cues.

\begin{table*}[t]
\centering
\caption{Association-module pair features. Each scalar is paired with an availability indicator, giving a 32-dimensional input.}
\label{tab:supp_features}
\small
\setlength{\tabcolsep}{5pt}
\renewcommand{\arraystretch}{1.1}
\begin{tabular}{@{}r p{0.24\textwidth} p{0.66\textwidth}@{}}
\toprule
\# & Feature & Definition \\
\midrule
1 & Appearance cosine & Cosine similarity between the mask-pooled features of the observation and the bank entry. \\
2 & Appearance margin & Candidate similarity minus the best competing-candidate similarity. \\
3 & Mask IoU & Overlap with the dormant identity's last mask, when available. \\
4 & Box IoU & IoU between the observation box and the candidate's last observed box. \\
5 & Center distance & Euclidean center displacement, normalized by the image diagonal. \\
6 & Log area ratio & Absolute log ratio of observation and candidate box areas. \\
7 & Log aspect-ratio difference & Absolute log difference between box aspect ratios. \\
8 & Motion residual & Center distance from the constant-velocity prediction in normalized image coordinates, divided by $\sqrt{2}$. \\
9 & Time gap & Time since the candidate was last observed, divided by the 30\,s retention limit. \\
10 & Detection confidence & Confidence of the observation. \\
11 & Tracker confidence & Most recent confidence stored for the dormant identity. \\
12 & Occlusion fraction & Fraction of the preceding 30 frames in which the identity was unobserved. \\
13 & Track age & Identity age, normalized and capped at 300\,s. \\
14 & Column margin & Candidate margin across observations competing for the same identity. \\
15 & Reciprocal best & Indicator that the observation and the identity are mutual best matches. \\
16 & Competitor count & Number of observations competing for the candidate identity. \\
\bottomrule
\end{tabular}
\end{table*}

The pair network produces one logit per eligible candidate:
\begin{align}
  h_{ik}^{(1)} &= \operatorname{GELU}\!\left(\operatorname{LN}(W_1 x_{ik} + b_1)\right), \\
  h_{ik}^{(2)} &= \operatorname{GELU}\!\left(W_2 \operatorname{Drop}_{0.1}(h_{ik}^{(1)}) + b_2\right), \\
  s_{ik} &= W_3 h_{ik}^{(2)} + b_3,
\end{align}
with widths $32 \rightarrow 128 \rightarrow 64 \rightarrow 1$. The \textsc{New} head receives the observation-level context
\begin{equation}
  c_i = [p_i, K_i, s_i^{(1)}, s_i^{(2)}, r_i^{\min}] \in \mathbb{R}^5,
\end{equation}
where $p_i$ is the detection confidence, $K_i$ the number of candidates, $s_i^{(1)}$ and $s_i^{(2)}$ the two largest pair logits (using $-10$ for each unavailable top-ranked logit), and $r_i^{\min}$ the smallest motion residual. Its logit is
\begin{equation}
  s_i^{\mathrm{new}} = (w_2^{\mathrm{new}})^\top \operatorname{GELU}(W_1^{\mathrm{new}} c_i + b_1^{\mathrm{new}}) + b_2^{\mathrm{new}},
\end{equation}
with hidden width 32. The association module has 13,026 trainable parameters and requires $12{,}352K_i+192$ weight multiply--accumulate operations for an observation with $K_i$ candidates.

The candidate and \textsc{New} logits form one softmax,
\begin{equation}
  P(y_i = j \mid o_i, \mathcal{B}_t) = \frac{\exp(s_{ij})}{\exp(s_i^{\mathrm{new}}) + \sum_{k=1}^{K_i} \exp(s_{ik})},
\end{equation}
with $j \in \{1, \ldots, K_i, \mathrm{new}\}$, trained with the cross-entropy loss
\begin{equation}
  \mathcal{L}_{\mathrm{assoc}} = -\frac{1}{B} \sum_{i=1}^B \log P(y_i^* \mid o_i, \mathcal{B}_t).
\end{equation}
The pair network and \textsc{New} head are trained jointly under this objective. At inference, Hungarian assignment enforces one-to-one candidate matching, and an assigned identity is accepted only if its pair logit is at least the observation's \textsc{New} logit.

\subsection{Accepted Observations and Duplicate Merge}
\label{sec:supp_merge}

\paragraph{Accepted observations:} We filter observations by detection confidence before identity association. For Crystals, all compared configurations use a confidence threshold of $0.23$, following CountVid for a matched comparison; the threshold is $\tau=0.5$ on the other benchmarks. An accepted observation must also have nonempty mask support after resizing to the $72\times72$ backbone feature grid, allowing its appearance feature to be pooled. Observations that fail either condition receive no identity assignment in that frame.

\paragraph{Duplicate merge:} After the video, we merge identities with at least five shared frames and mean mask IoU $>0.5$. We also merge short handovers with at most two shared frames, endpoint gap $\leq 5$ frames, and mask-center distance $d_c < 50$ native-resolution pixels. For crowded queries, we apply a stricter handover rule to avoid merging neighboring objects. Specifically, tracks are linked only if they have no shared frames, a gap
$g \in [1,5]$, an area ratio
$A_{\max}/A_{\min} < 2$, and a motion residual
$d < 0.25h_{\mathrm{prev}}$. All handover links are enforced one-to-one, and any resulting chain retains the earliest identity.

\paragraph{Long videos:} Videos too large for one tracking session in GPU memory are processed in overlapping sessions; for MOT20-Count, we needed this only for MOT20-05; MOT20-01 and MOT20-02 run in a single session. The overlapping frames warm up the tracker; the identity bank is advanced over them so that identities from the previous session are candidates for re-detections in the next one, and tracks are linked one-to-one across the boundary when their masks overlap with a mean IoU of at least 0.5 over at least three frames.

\subsection{Trajectory-Level Verifier}
\label{sec:supp_verifier}

Let $Z$ denote the merged identity set in the post-tracking formulation, and let $Z^{\prime}\subseteq Z$ be the set retained by verification; the final count is $\widehat{N}_q=|Z^{\prime}|$. After the duplicate merge, a trajectory-level verifier removes identities that are spurious or duplicates of another identity. It scores each identity on 18 features computed from its own trajectory, the other identities in the same query, and the association decision that created it (Table~\ref{tab:supp_verifier_features}). The creation-related features are detection confidence, the highest candidate appearance similarity, and candidate count (features 13--15). The verifier receives no category name.

\begin{table*}[t]
\centering
\caption{Verifier features. Overlap features compare an identity's boxes with the boxes of the other identities of the same query on the same frames.}
\label{tab:supp_verifier_features}
\small
\setlength{\tabcolsep}{5pt}
\renewcommand{\arraystretch}{1.1}
\begin{tabular}{@{}r p{0.24\textwidth} p{0.66\textwidth}@{}}
\toprule
\# & Feature & Definition \\
\midrule
1 & Log observation count & Logarithm of the number of frames with an accepted observation of the identity. \\
2 & Span & Time between the first and last observation, in seconds. \\
3 & Density & $n_k/(\Delta t_k r+1)$, where $n_k$ is the observation count, $\Delta t_k$ the span in seconds, and $r$ the processing rate. \\
4 & Relative area & Median box area divided by the median over all identities of the query. \\
5 & Aspect ratio & Median box height divided by median box width. \\
6 & Maximum overlap & Largest IoU with another identity's box over the lifetime. \\
7 & Overlap fraction & Share of its frames with an IoU above 0.5 with another identity. \\
8 & Maximum containment & Largest share of its box that lies inside another identity's box. \\
9 & Containment fraction & Share of its frames in which more than 80\% of its box lies inside another identity's box. \\
10 & Overlapping identities & Number of other identities with an IoU above 0.3 on at least one shared frame. \\
11 & Log identity count & Logarithm of the number of identities of the query. \\
12 & Length rank & Rank of its observation count among the query's identities, normalized to $(0, 1]$. \\
13 & Creation confidence & Detection confidence when the identity was created ($-1$ if no decision was logged). \\
14 & Creation similarity & Highest appearance cosine to a dormant candidate at creation (0 if none). \\
15 & Creation candidates & Number of eligible dormant candidates at creation. \\
16 & Reactivations & Number of times the identity was reactivated. \\
17 & Start position & First observation, relative to the time extent of the query. \\
18 & End position & Last observation, relative to the time extent of the query. \\
\bottomrule
\end{tabular}
\end{table*}

\paragraph{Labels:} Training labels come from ground-truth tracks on annotated frames. On every annotated frame within an identity's lifetime, we match its box to the ground-truth object it overlaps with an IoU of at least 0.5, and assign the identity to the object it matches most often. Among the identities assigned to one object, the one that matches it most often is \emph{good}, and all others are \emph{duplicates}. An identity that matches no object is labeled \emph{spurious}. We use this label only for categories that are exhaustively labeled. For datasets with amodal ground-truth boxes (DanceTrack and MOT17), a match may also satisfy an IoU of at least 0.3 with at least 80\% of the predicted box inside the ground-truth box, so that a partially occluded person is not labeled spurious. Identities with no annotated frame in their lifetime are not used.

\paragraph{Training data:} We train the verifier on the identities the frozen IDTCount pipeline produces on OVIS-valid, DanceTrack-train, and MOT17-train at 3\,fps, none of which is an evaluation set. Together, these give 1,494 good, 393 duplicate, and 131 spurious identities. OVIS-valid contributes 746, 63, and 78; DanceTrack-train 408, 320, and 10; and MOT17-train 340, 10, and 43. Videos are assigned deterministically to fitting or selection using a hash of the source and video identifier, with all queries from a video kept together. The resulting split contains 1,610 labeled identities for fitting and 408 labeled identities across 63 selection queries.

\paragraph{Model and threshold:} The classifier is a histogram gradient-boosting classifier (depth 3, 200 boosting iterations, learning rate 0.05, L2 regularization 1.0) trained to predict whether an identity is not good. We remove an identity if its predicted probability meets the threshold, and the retained identities form the count. We select the removal threshold by minimizing count MAE over the 63 held-out queries, searching $\{0.05,0.10,\ldots,0.95\}$ together with a no-removal option and breaking ties toward the higher threshold. On this model-selection split, the count MAE falls from 1.889 without removal to 0.619 at a threshold of 0.5, and the held-out AUC is 0.957. A variant without the five overlap and containment features (features 6--10) reaches an AUC of 0.915 and an MAE of 0.873 at its own best threshold, so we keep all 18 features. We keep the model, feature set, and threshold of 0.5 fixed wherever the verifier is applied.

\section{Further Implementation Details}
\label{sec:supp_training}

\subsection{Training Data}

Each training event contains one observation that needs a decision, its candidate set, the 32-dimensional pair features, and the target. Events are generated by replaying training videos through the identity layer and labeling decisions against ground-truth tracks. The target is \textsc{Match} for an available corresponding dormant identity and \textsc{New} for the first labeled decision of an identity or an unmatched detection.  Events before confirmation, and events of tracks that never reach two matches, are ignored. The first decision event at or after confirmation establishes the identity in the supervision and is labeled \textsc{New}. A native-track continuation produces no decision event, so this need not occur on the second matched frame. Subsequent decisions whose corresponding identity is absent from the candidate set are excluded from supervised training. These rules affect labels only. At inference, an observation with no eligible candidate creates a new identity. In CityFlowV2, whose events are built from ground-truth boxes, a track's first observation is \textsc{New} and every later observation matches its own prior state. CityFlowV2 events are built independently for each camera. Track identities are scoped by scenario and camera, so every candidate comes from the same camera as the observation and cross-camera identity is never used. CityFlowV2 therefore broadens training with same-camera reactivation examples involving dense vehicle interactions, spatial competition, motion change, and re-entry; it does not teach matching across cameras or viewpoints and is used for training only. The module never receives category names or identifiers.

Table~\ref{tab:supp_h2data} reports the raw size of each source and the events usable for the candidate-or-\textsc{New} objective. The VIS sources contribute 608 \textsc{Match} events among 7,493 usable events; adding CityFlowV2 increases these totals to 24,069 and 31,414, respectively. Training minibatches contain equal shares of \textsc{Match} and \textsc{New} events, with an unweighted cross-entropy loss. Selection data retain their natural composition.

\begin{table}[t]
\centering
\caption{Association-module training and selection events. Trainable: events usable for the candidate-or-\textsc{New} objective. Parentheses: trainable selection events.}
\label{tab:supp_h2data}
\small
\setlength{\tabcolsep}{4pt}
\renewcommand{\arraystretch}{1.1}
\begin{tabular}{@{}lrrr@{}}
\toprule
Source & Raw train & Trainable & Selection \\
\midrule
OVIS & 7,213 & 4,038 & 1,389 (692) \\
YouTube-VIS & 7,456 & 2,193 & 227 (94) \\
LV-VIS & 7,935 & 1,262 & 65 (4) \\
CityFlowV2 & 23,921 & 23,921 & 7,754 (7,754) \\
\midrule
Total & 46,525 & 31,414 & 9,435 (8,544) \\
\bottomrule
\end{tabular}
\end{table}

\begin{table*}[!h]
\centering
\caption{Data separation of the association module. Videos are identified by split and identifier (ev.: events).}
\label{tab:supp_separation}
\footnotesize
\setlength{\tabcolsep}{4pt}
\renewcommand{\arraystretch}{1.1}
\begin{tabular}{@{}l p{0.25\textwidth} p{0.24\textwidth} p{0.3\textwidth} l@{}}
\toprule
Source & Training & Selection & Evaluation & Eval. overlap \\
\midrule
OVIS & train split, 607 videos (7,213 ev.) & valid split, 140 videos (1,389 ev.) & test split, 154 videos, 224 queries & none \\
YT-VIS & train split, 1,662 videos (7,456 ev.) & 30 train-split videos (227 ev.) & valid split, 308 videos, 481 queries & none \\
LV-VIS & train split, 2,575 videos (7,935 ev.) & 26 train-split videos (65 ev.) & val split, 679 videos, 2,462 queries & none \\
CityFlowV2 & S01 and S04, 30 cameras (23,921 ev.) & S03, 5 cameras (7,754 ev.) & not evaluated & -- \\
\bottomrule
\end{tabular}
\end{table*}

\subsection{Optimization and Checkpoint Selection}
\label{sec:supp_selection}

We train the association module with AdamW (learning rate $10^{-3}$, weight decay $10^{-4}$, batch size 32) for at most 100 epochs, with early stopping after 15 epochs without improvement. After each epoch, we measure the reactivation accuracy on each selection source over its solvable \textsc{Match} events,
\begin{equation}
  \operatorname{AssignAcc}_d = \frac{1}{|\mathcal{M}_d|} \sum_{e \in \mathcal{M}_d} \mathbf{1}[\widehat{y}_e = y_e^*],
\end{equation}
where both a wrong candidate and \textsc{New} count as errors. Among the epochs that keep wrong-candidate selection in \textsc{Match} events and false reactivation in \textsc{New} events below per-source ceilings, we keep the epoch that maximizes $\min_d \operatorname{AssignAcc}_d$.


\subsection{Data Separation}
\label{sec:supp_separation}

Table~\ref{tab:supp_separation} summarizes the training, selection, and evaluation videos for each source. We check overlap using both split and video identifier, since LV-VIS reuses numeric identifiers across splits. OVIS selection uses the disjoint validation split, while CityFlowV2 selection uses scenario S03, separate from training scenarios S01 and S04. YT-VIS and LV-VIS selection are not held out: all 30 and 26 selected videos also occur in fitting, contributing 227 and 65 events to their respective training totals of 7,456 and 7,935. At the selected epoch, the disjoint OVIS selection set gives the lowest reactivation accuracy, 0.2125 over 80 \textsc{Match} events; YT-VIS and LV-VIS achieve 1.0 over 9 and 1 event, respectively. Their safety-ceiling checks therefore include events seen during fitting. No evaluation video appears in training or selection. The scored \textsc{Match}/\textsc{New} events are 80/612 for OVIS validation, 9/85 for YT-VIS, 1/3 for LV-VIS, and 7,676/78 for CityFlowV2 S03.

\subsection{Evaluation Protocol}
\label{sec:supp_protocol}

\paragraph{Compared systems:} CountVid is our own run of its released text-only configuration on the same video-query pairs as IDTCount on BURST-Count, TAO-Count, and the three identity benchmarks. On MOT20-Count, we use the MAE and RMSE reported in its paper; on Penguins and AnimalTrack-Count, we run its released text-only configuration. ByteTrack and MASA associate the detections of the same frozen SAM3 detector as IDTCount. The SAM3 baseline runs as released; Table~\ref{tab:supp_eviction} separately evaluates the effect of tracker memory eviction. SAM3Count results are reproduced with its released inference code and SAM3 checkpoint in balanced mode.

\paragraph{Identity-layer transfer:} For ByteTrack and MASA, ``+IDTCount'' adds its identity layer components but replaces SAM3's tracker with these two trackers. These external trackers supply native-track continuity on SAM3 detections, while the identity layer pools 256-dimensional appearance features from the frozen SAM3 image backbone inside each detection mask. These experiments then evaluate the identity layer of IDTCount as a component.

\paragraph{Identity metrics:} We compute HOTA, AssA, and DetA using TrackEval, averaging over localization thresholds $\alpha\in\{0.05,0.10,\ldots,0.95\}$~\cite{luiten2021hota}, and report IDF1~\cite{Ristani_2016_IDF1} and ID switches. A ground-truth track is fragmented if it is matched to more than one predicted identity over its frames. Long-gap recovery is the fraction of objects that return with the same identity after being missed for at least 1\,s.

\paragraph{Counting metrics:} For $N$ video-query pairs with error $e_i = \widehat{N}_i - N_i$,
\begin{align}
  \operatorname{MAE} &= \frac{1}{N} \sum_i |e_i|, \\
  \operatorname{RMSE} &= \sqrt{\frac{1}{N} \sum_i e_i^2}, \\
  \operatorname{Exact} &= \frac{1}{N} \sum_i \mathbf{1}[e_i = 0], \\
  \operatorname{Close} &= \frac{1}{N} \sum_i \mathbf{1}\!\left[|e_i| \le 1 + \lfloor 0.05\, N_i \rfloor\right].
\end{align}
Overcount and undercount rates are the fractions with $e_i > 0$ and $e_i < 0$. ``Close'' is a count-dependent tolerance, not strictly ``within one''.

\section{Further Dataset Details}
\label{sec:supp_data}
\subsection{Benchmark Statistics}

Table~\ref{tab:datasets} summarizes the evaluation benchmarks and the data used to train the association module and verifier. Association-selection sources are listed separately in Table~\ref{tab:supp_separation}, and we describe the BURST-train development subset used for sensitivity analysis in Sec.~\ref{sec:supp_retention}.

\begin{table}[t]
\centering
\caption{Data sources. A query is a video-category pair. Objects: range of the ground-truth count per query. Frames: fewest and most frames per video at the listed frame rate. A dash marks a statistic we do not report for that source.}
\label{tab:datasets}
\small
\setlength{\tabcolsep}{3pt}
\renewcommand{\arraystretch}{1.08}
\resizebox{\columnwidth}{!}{%
\begin{tabular}{@{}lrrrrrr@{}}
\toprule
Data source & Videos & Queries & Classes & Objects & Frames & FPS \\
\midrule
\multicolumn{7}{@{}l}{\textit{Counting}} \\
BURST-Count~\cite{Athar_2023_BURST} & 868 & 1,556 & 274 & 1--10 & 37--235 & 3 \\
TAO-Count~\cite{dave2020tao,amini2025countvid} & 357 & 639 & 139 & 1--10 & 24--235 & 3 \\
MOT20-Count~\cite{dendorfer2020mot20,amini2025countvid} & 3 & 3 & 1 & 80--1,203 & 52--398 & 3 \\
AnimalTrack-Count~\cite{zhang2023animaltrack} & 25 & 25 & 10 & 8--134 & 196--2,126 & 30 \\
Penguins~\cite{amini2025countvid} & 3 & 3 & 1 & 28--59 & 34--94 & 3 \\
Crystals~\cite{amini2025countvid} & 7 & 7 & 1 & 8--154 & 24--46 & $\sim$6--7 \\
\midrule
\multicolumn{7}{@{}l}{\textit{Identity}} \\
OVIS-test~\cite{qi2022occluded} & 154 & 224 & 25 & 1--19 & 21--309 & 5 \\
YT-VIS~\cite{Yang_2019_YTVIS} & 308 & 481 & 40 & 1--13 & 6--84 & 6 \\
LV-VIS~\cite{wang2023towards} & 679 & 2,462 & 621 & 1--32 & 5--54 & 5 \\
\midrule
\multicolumn{7}{@{}l}{\textit{Association-module training (train splits)}} \\
OVIS-train~\cite{qi2022occluded} & 607 & 886 & 25 & 1--44 & 15--500 & 5 \\
YT-VIS-train~\cite{Yang_2019_YTVIS} & 1,662 & -- & -- & -- & -- & 6 \\
LV-VIS-train~\cite{wang2023towards} & 2,575 & -- & -- & -- & -- & 5 \\
CityFlowV2 S01, S04~\cite{Tang_2019_CVPR,Naphade_2022_CVPR} & 30 cameras & -- & -- & -- & -- & -- \\
\midrule
\multicolumn{7}{@{}l}{\textit{Verifier training}} \\
OVIS-valid~\cite{qi2022occluded} & 140 & 198 & 25 & 1--15 & 9--176 & 3 \\
DanceTrack-train~\cite{sun2022dancetrack} & 40 & 40 & 1 & 4--40 & 61--325 & 3 \\
MOT17-train~\cite{dendorfer2021motchallenge} & 7 & 7 & 1 & 27--133 & 53--180 & 3 \\
\bottomrule
\end{tabular}
}
\end{table}
\subsection{BURST-Count}

BURST~\cite{Athar_2023_BURST} inherits TAO's federated labeling~\cite{dave2020tao}, so real instances of those categories can be missing from the ground truth. A count is only meaningful when every instance is labeled. We therefore build BURST-Count from the BURST test split by keeping only categories that are exhaustively labeled, following the approach in~\cite{amini2025countvid} for TAO-Count. This gives 1,556 video-query pairs from 868 videos and 274 categories, with 1--10 objects per query. Unlike TAO-Count, BURST-Count keeps the AVA and HACS videos. The category names serve as text prompts.

\paragraph{TAO-Count:} TAO-Count's 639 queries are TAO's validation queries with exhaustively labeled categories outside the AVA and HACS videos, except for one Charades query; all 639 are flagged as exhaustive in BURST validation, and the ground-truth counts agree with BURST on 638 of them.

\paragraph{AnimalTrack-Count and Penguins:} AnimalTrack-Count uses the 25 test videos of AnimalTrack~\cite{zhang2023animaltrack}, each showing one animal species per video, with every animal of that species labeled on every frame. The prompt is the species name, and the ground truth is the number of distinct labeled tracks. Penguins are part of the Science-Count benchmark of CountVid~\cite{amini2025countvid}: three videos queried with the prompt ``penguin''.

\subsection{Crystals and Detector Transfer}
\label{sec:supp_crystals}
Crystals contains seven videos queried with ``white crystal''. Crystals provides only frame-level and overall video counts, with no ground-truth box or mask annotations. We process all supplied evaluation frames, giving 24--46 processed frames per video in the completed runs. The source specifies an approximate rate of 6--7\,fps without exact timestamps; IDTCount uses a nominal rate of 6.5\,fps for its temporal features.

We observed that the detectors struggle to localize objects in this dataset when prompted with text alone, so we used internal exemplars. Internal exemplars are the three highest-confidence valid boxes selected from a video's text-only detection pass. This selection requires no ground-truth exemplar. The selected boxes and their source images then provide visual conditioning for the exemplar-enabled pass. Specifically, CountGD-Box reuses cached source-region tokens, whereas the SAM3 adapter combines source visual tokens with native target geometry. Unlike causal first-frame prompting, this procedure selects exemplars across the entire video.

The native CountVid configuration retains its temporal filter with width three and SAM2 tracking. Its Crystal runs use image batches of two, object batches of ten, and filter batches of ten. CountVid + IDTCount disables the temporal filter and adds the dormant identity bank, learned association, and duplicate merge, without trajectory verification, while retaining the detector and SAM2 tracker. Its 256-dimensional appearance features are pooled inside the SAM2 output masks from an additional frozen SAM3 image-backbone pass on each processed frame; they are not extracted from SAM2. CountGD-Box + SAM3 + IDTCount instead supplies CountGD-Box detections to SAM3 tracking and applies the same bank, association, and duplicate merge without trajectory verification. These experiments evaluate transferring the identity layer to other systems and what happens when the detector is replaced. Table~\ref{tab:supp_crystals_counts} gives their per-video predictions.

\begin{figure*}[t]
\centering
\includegraphics[width=\textwidth]{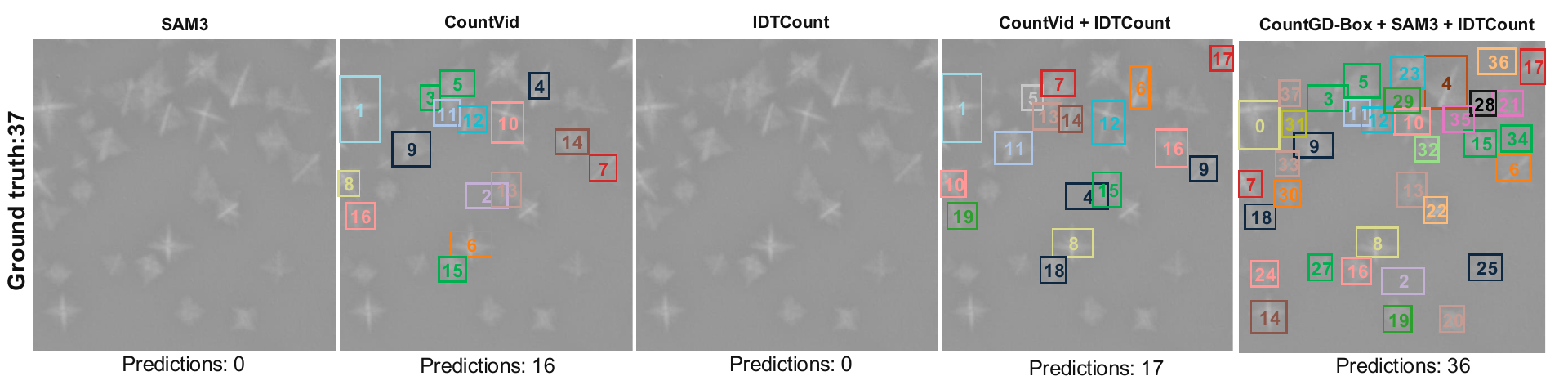}
\caption{Crystals comparison on one frame with the prompt ``white crystal'' and internal exemplars (GT count: 37). Only count-level ground truth is available because no boxes or masks are released.}
\label{fig:supp_crystals_overlays}
\end{figure*}

Figure~\ref{fig:supp_crystals_overlays} compares five configurations on one frame under the text-plus-internal-exemplar setting. SAM3 as a baseline and IDTCount (SAM3 base tracker + identity layer) with the SAM3 detector both produce zero observations, showing that the identity layer cannot recover objects that the detector fails to localize. With CountVid detections, adding IDTCount as an identity layer changes the frame total from 16 to 17. CountGD-Box detections with SAM3 tracking and IDTCount association yield 36 against the frame count of 37, demonstrating the contribution of detector coverage in this example.

\section{Further Quantitative Results}
\label{sec:supp_results}

\subsection{Identity and Frame-Level Count Errors}
\label{sec:supp_identity_errors}
Table~\ref{tab:supp_identity_errors} reports additional metrics for the same CountVid and final IDTCount predictions as the main-paper identity table. Count MAE and RMSE compare the predicted and annotated numbers of visible objects on each evaluated query-frame. Exact accuracy is the percentage of query-frames with equal counts. Over and under are the percentages with positive and negative count error. These frame-level measures are distinct from final video-level unique-instance counts. The number of fragmented ground-truth tracks (Frag.) counts tracks matched to more than one predicted identity over time. Dup. is the percentage of annotated object-frame opportunities exhibiting duplicate predictions, using the evaluator's IoU matching threshold of 0.5.

\begin{table*}[t]
\centering
\caption{Frame-level counting and identity errors of CountVid and IDTCount on the VIS benchmarks. Best per dataset is bold.}
\label{tab:supp_identity_errors}
\small
\setlength{\tabcolsep}{4pt}
\resizebox{\textwidth}{!}{%
\begin{tabular}{@{}llrrrrrrr@{}}
\toprule
Dataset & Method & Count MAE $\downarrow$ & Count RMSE $\downarrow$ & Exact (\%) $\uparrow$ & Over (\%) $\downarrow$ & Under (\%) $\downarrow$ & Frag. $\downarrow$ & Dup. (\%) $\downarrow$ \\
\midrule
OVIS-test & CountVid & 1.008 & 1.987 & 54.18 & 15.18 & 30.64 & 123 & 3.62 \\
OVIS-test & IDTCount & \textbf{0.394} & \textbf{0.816} & \textbf{71.06} & \textbf{13.39} & \textbf{15.55} & \textbf{67} & \textbf{3.21} \\
\midrule
YT-VIS & CountVid & 1.179 & 4.460 & 60.32 & \textbf{19.54} & 20.14 & 64 & 2.21 \\
YT-VIS & IDTCount & \textbf{0.806} & \textbf{3.580} & \textbf{73.32} & 19.59 & \textbf{7.09} & \textbf{51} & \textbf{1.90} \\
\midrule
LV-VIS & CountVid & 1.130 & 2.809 & 33.12 & 18.81 & 48.07 & 47 & 0.91 \\
LV-VIS & IDTCount & \textbf{0.421} & \textbf{1.419} & \textbf{74.04} & \textbf{10.74} & \textbf{15.23} & \textbf{36} & \textbf{0.45} \\
\bottomrule
\end{tabular}%
}
\end{table*}
Compared with CountVid, IDTCount lowers frame-level count MAE and RMSE and increases exact accuracy on all three benchmarks. It also reduces undercounting, fragmented tracks, and duplicate events throughout. Overcounting decreases on OVIS-test and LV-VIS and is nearly unchanged on YT-VIS (19.59\% versus 19.54\%).

\subsection{Dense Scenes and Long Videos}

We run inference on an NVIDIA H100 GPU with 80\,GB of memory. Dense scenes and long videos run in a single tracking session when they fit within this capacity; otherwise, we use overlapping sessions to keep GPU memory within the hardware limit. Table~\ref{tab:supp_mot20} gives the MOT20-Count results per video. MOT20-01 and MOT20-02 run in a single tracking session. MOT20-05 exceeds the H100's 80\,GB capacity in a single session and therefore uses overlapping sessions, with a peak memory of 36.5\,GB.

\begin{table}[!h]
\centering
\caption{MOT20-Count per-video counts and peak GPU memory at 3\,fps.}
\label{tab:supp_mot20}
\small
\setlength{\tabcolsep}{3pt}
\renewcommand{\arraystretch}{1.08}
\resizebox{\columnwidth}{!}{%
\begin{tabular}{@{}lrrrr@{}}
\toprule
Video & Frames & Peak GiB & Ground Truth & IDTCount \\
\midrule
MOT20-01 & 52 & 18.7 & 80 & 78 \\
MOT20-02 & 334 & 45.7 & 278 & 260 \\
MOT20-05 & 398 & 36.5 & 1,203 & 1,216 \\
\bottomrule
\end{tabular}
}
\end{table}

\subsection{Component Counting Ablation}
Table~\ref{tab:supp_components_counting} shows the identity-quality ablation with counting errors on BURST-Count. The SAM3 baseline uses the released tracker with eviction; all rows with the association module use eviction, as in IDTCount. The first transition therefore changes both association and the eviction setting; Table~\ref{tab:supp_eviction} isolates the eviction tradeoff.

\begin{table}[!h]
\centering
\caption{Cumulative component counting ablation on BURST-Count (1,556 queries). MAE with 95\% intervals from a video-clustered bootstrap (2,000 resamples). Over and under are the numbers of queries with positive and negative error.}
\label{tab:supp_components_counting}
\small
\setlength{\tabcolsep}{3pt}
\renewcommand{\arraystretch}{1.08}
\resizebox{\columnwidth}{!}{%
\begin{tabular}{@{}lcc@{}}
\toprule
 & \multicolumn{2}{c}{BURST-Count} \\
\cmidrule(lr){2-3}
Variant & MAE [95\% CI] $\downarrow$ & Over / under \\
\midrule
SAM3 & 0.656 [0.572, 0.750] & 426 / \textbf{150} \\
\quad + association module & 0.651 [0.572, 0.747] & 423 / 153 \\
\quad + merge & 0.643 [0.567, 0.735] & 421 / 154 \\
\quad + verifier & \textbf{0.537} [0.477, 0.605] & \textbf{373} / 166 \\
\bottomrule
\end{tabular}
}
\end{table}

\begin{table*}[t]
\centering
\caption{Crystals per-video predicted counts. CountVid + IDTCount replaces CountVid's temporal filter while retaining its detector and SAM2 tracker. CountGD-Box + SAM3 + IDTCount supplies CountGD-Box detections to SAM3 tracking. Both extensions evaluate identity-layer transfer.}
\label{tab:supp_crystals_counts}
\small
\setlength{\tabcolsep}{5pt}
\renewcommand{\arraystretch}{1.1}
\resizebox{\textwidth}{!}{%
\begin{tabular}{@{}lrrrrrrr@{}}
\toprule
Method / Video
& ma2035\_023 & ma2035\_024 & ma2035\_039
& ma2035\_040 & ma2035\_069 & ma2035\_072
& ma2035\_169 \\
\midrule
Ground truth & 111 & 154 & 27 & 37 & 109 & 77 & 8 \\
\midrule
\multicolumn{8}{@{}l}{\textit{Text only}} \\
\midrule
SAM3 & 0 & 0 & 0 & 0 & 0 & 0 & 0 \\
CountVid & 1 & 1 & 11 & 9 & 11 & 2 & \textbf{11} \\
CountVid + IDTCount & 1 & 1 & 17 & 21 & 23 & 5 & 15 \\
CountGD-Box + SAM3 + IDTCount & 1 & 1 & 17 & 21 & 24 & 5 & 15 \\
\midrule
\multicolumn{8}{@{}l}{\textit{Text + internal exemplars}} \\
\midrule
SAM3 & 0 & 0 & 0 & 0 & 0 & 0 & 0 \\
CountVid & 96 & 41 & \textbf{27} & 16 & 90 & 59 & 17 \\
CountVid + IDTCount & \textbf{105} & 65 & 3 & 18 & 95 & 71 & 17 \\
CountGD-Box + SAM3 + IDTCount & \textbf{105} & \textbf{71} & 33 & \textbf{47} & \textbf{105} & \textbf{74} & 17 \\
\bottomrule
\end{tabular}%
}
\end{table*}

\subsection{Tracker Memory Eviction}
\label{sec:supp_eviction}

SAM3's tracker memory grows with every processed frame, so IDTCount evicts frames that have left its attention window. Table~\ref{tab:supp_eviction} compares SAM3 with and without eviction. Counting metrics are MAE / RMSE, and latency and memory are measured over the 224 OVIS-test queries using the protocol in Sec.~\ref{sec:supp_efficiency}. Eviction adds 12.8\,ms of tracking latency per frame while reducing peak GPU memory from 6.88 to 6.73\,GiB on OVIS-test. It changes MAE by only 0.002 on each counting benchmark, so we use it to bound tracker-memory growth. The full IDTCount system improves counting accuracy over released SAM3 while running at 105.2\,ms per frame, compared with 95.3\,ms per frame for SAM3; verification is timed separately.

\begin{table}[!h]
\centering
\caption{Effect of tracker memory eviction on SAM3 counting error (MAE / RMSE), latency, and GPU memory.}
\label{tab:supp_eviction}
\small
\setlength{\tabcolsep}{4pt}
\renewcommand{\arraystretch}{1.08}
\resizebox{\columnwidth}{!}{%
\begin{tabular}{@{}lcccc@{}}
\toprule
SAM3 & BURST-Count & TAO-Count & ms/frame & Peak GiB \\
\midrule
As released & \textbf{0.654} / 1.792 & 0.653 / 1.606 & \textbf{95.3} & 6.88 \\
With eviction & 0.656 / \textbf{1.782} & \textbf{0.651} / \textbf{1.603} & 108.1 & \textbf{6.73} \\
\bottomrule
\end{tabular}
}
\end{table}

\subsection{Retention Window and Lost-Track Gate}
\label{sec:supp_retention}
We vary the retention window and lost-track gate separately on 461 exhaustively labeled BURST-train queries from 278 videos, processed at 3\,fps, holding all other settings fixed and evaluating without trajectory verification. These are development data, disjoint from BURST-Count evaluation. The reference configuration uses a 30\,s retention window and a 0.167\,s lost-track gate. Table~\ref{tab:supp_retention} reports MAE, RMSE, and the MAE difference from the reference, with paired video-clustered 95\% bootstrap intervals from 2,000 resamples.

\begin{table}[!h]
\centering
\caption{Dormant-bank sensitivity on exhaustive BURST-train queries. $\Delta$ is the MAE difference from the reference.}
\label{tab:supp_retention}
\small
\setlength{\tabcolsep}{3pt}
\renewcommand{\arraystretch}{1.08}
\resizebox{\columnwidth}{!}{%
\begin{tabular}{@{}llccc@{}}
\toprule
Setting & Value & MAE $\downarrow$ & RMSE $\downarrow$ & $\Delta$ [95\% CI] \\
\midrule
Retention window & 5\,s & 0.3861 & 0.901 & +0.0043 [0.000, +0.011] \\
 & 10\,s & 0.3861 & 0.901 & +0.0043 [0.000, +0.011] \\
 & 15\,s & 0.3796 & 0.885 & $-$0.0022 [$-$0.010, +0.004] \\
 & 30\,s (reference) & 0.3818 & 0.891 & 0 (reference) \\
\midrule
Lost-track gate & 0.167\,s (reference) & 0.3818 & 0.891 & 0 (reference) \\
 & 0.67\,s & 0.3839 & 0.900 & +0.0022 [0.000, +0.007] \\
 & 1.0\,s & 0.3839 & 0.900 & +0.0022 [0.000, +0.007] \\
 & 2.0\,s & 0.3839 & 0.900 & +0.0022 [0.000, +0.007] \\
\bottomrule
\end{tabular}%
}
\end{table}
Counting error changes little across the tested retention windows and lost-track gates, and all paired difference intervals include zero. We therefore use a 30\,s retention window and a 0.167\,s lost-track gate, which provide recovery across extended absences without retaining stale candidates indefinitely.

\subsection{Efficiency, Runtime, and Memory}
\label{sec:supp_efficiency}

Table~\ref{tab:latency} reports per-frame latency, the median time per frame of the tracking loop averaged over the 224 OVIS-test queries; for CountVid, whose pipeline is not streaming, we measure inference time per frame. For IDTCount, this measures the frame loop, including identity-bank updates and learned association. Post-video duplicate merging and verification are excluded from this per-frame latency; verifier runtime is reported below. Memory is the mean per-query peak allocated GPU memory in GiB, using the same units as Table~\ref{tab:supp_eviction}. Figure~\ref{fig:analysis_pair} uses its profiler summary in decimal GB; the small numerical difference reflects the reporting convention. SAM3 is reported as released, without memory eviction; IDTCount uses eviction.

\begin{table}[!h]
\centering
\caption{Per-frame latency and peak GPU memory on 224 OVIS-test queries.}
\label{tab:latency}
\small
\setlength{\tabcolsep}{6pt}
\renewcommand{\arraystretch}{1.08}
\begin{tabular}{@{}lcc@{}}
\toprule
Method & Latency (ms) $\downarrow$ & Peak GPU (GiB) $\downarrow$ \\
\midrule
CountVid~\cite{amini2025countvid} (AAAI '26) & 1,424.3 & 14.63 \\
SAM3~\cite{carion2026sam3} (ICLR '26) & \textbf{95.3} & 6.88 \\
SAM3Count~\cite{Owusu_2026_CVPR} & 123.5 & 6.91 \\
IDTCount (ours) & 105.2 & \textbf{6.73} \\
\bottomrule
\end{tabular}
\end{table}

\begin{figure*}[!h]
\centering
\includegraphics[width=\textwidth]{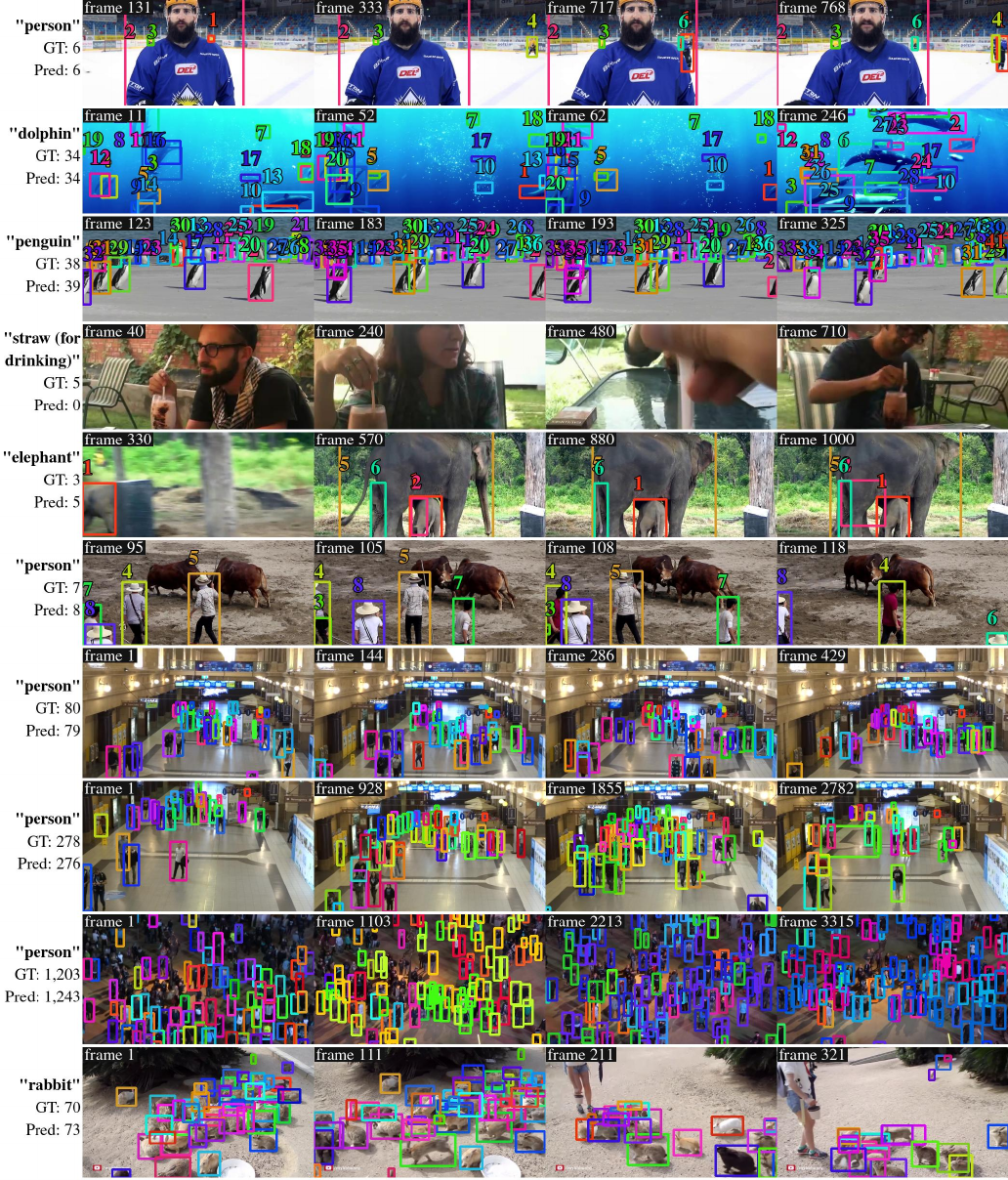}
\caption{IDTCount on ten videos, with four key frames per row. Labels show the text query, ground-truth count (GT), and predicted count (Pred). Rows 1--3 show accurate counts; rows 4--6 illustrate limitations; rows 7--10 show dense scenes with intermediate identity counts. Box numbers denote assigned identities and are omitted in dense scenes.}
\label{fig:supp_qualitative}
\end{figure*}

\paragraph{Verifier runtime:} Post-video verification takes 4.42\,s across the 224 OVIS-test queries, averaging 19.7\,ms per query. The CPU timer includes trajectory-feature computation and classifier prediction, excluding file loading and output writing. This cost is reported separately from tracking latency.

\section{Further Qualitative Results}
\label{sec:supp_qualitative}

Additional qualitative videos are provided in the accompanying ZIP submission.

Fig.~\ref{fig:supp_qualitative} shows IDTCount on ten videos, grouped into accurate counts, limitations, and dense scenes. Each row is one video with four key frames; the label beside the row gives the text query, the ground-truth count (GT), and the number of identities IDTCount assigns (Pred).

The first three rows (ice hockey, dolphin, and penguin) illustrate accurate counting with many interacting and overlapping objects.

The next three rows illustrate missed detections, overcounting, and incorrect identity reuse. The frozen detector misses every straw, leaving no observation for IDTCount to count. The elephant example shows overcounting. In the OVIS person example, an exiting person and a different entering person receive the same identity (ID~4).

The last four rows show dense scenes: the three MOT20-Count videos and an AnimalTrack-Count video of rabbits. For these four videos, Pred is the intermediate number of accumulated identities; the final counts for the three MOT20 videos are in Table~\ref{tab:supp_mot20}. The rabbit row illustrates dense-scene tracking before post-video refinement; its displayed Pred is not the final video-level counting result.

\end{document}